\documentclass[preprint,journal]{vgtc}       %

\ifpdf%
  \pdfoutput=1\relax                   %
  \usepackage{graphicx}                %
  \DeclareGraphicsExtensions{.pdf,.png,.jpg,.jpeg} %
\else%
  \usepackage{graphicx}                %
  \DeclareGraphicsExtensions{.eps}     %
\fi%

\graphicspath{{figures/}{pictures/}{images/}{./}} %

\DeclareMathAlphabet{\altmathcal}{OMS}{cmsy}{m}{n}

\usepackage{amsmath,amsfonts,bm}

\def\eqref#1{equation~\ref{#1}}

\def\1{\bm{1}}

\def\vg{{\bm{g}}}
\def\vh{{\bm{h}}}

\def\vx{{\bm{x}}}
\def\vy{{\bm{y}}}
\def\vz{{\bm{z}}}

\def\mW{{\bm{W}}}

\DeclareMathAlphabet{\mathsfit}{\encodingdefault}{\sfdefault}{m}{sl}
\SetMathAlphabet{\mathsfit}{bold}{\encodingdefault}{\sfdefault}{bx}{n}

\DeclareMathOperator*{\argmax}{arg\,max}

\usepackage{amsmath}
\usepackage{amsfonts}
\usepackage{microtype}                 %
\PassOptionsToPackage{warn}{textcomp}  %
\usepackage{textcomp}                  %
\usepackage{mathptmx}                  %
\usepackage{times}                     %
\usepackage{cite}                      %
\usepackage{tabu}                      %
\usepackage{booktabs}                  %
\usepackage{float}
\usepackage{soul}
\ieeedoi{10.1109/TVCG.2021.3114845}

\onlineid{0}

\vgtccategory{Research}
\vgtcpapertype{application/design study}

\title{GenNI: Human-AI Collaboration for Data-Backed Text Generation}

\author{Hendrik Strobelt, Jambay Kinley, Robert Krueger, Johanna Beyer, Hanspeter Pfister, Alexander M. Rush}

\authorfooter{
\item Hendrik Strobelt -  IBM Research \& MIT-IBM Watson AI Lab, Cambridge, Massachusetts, United States. Email: hendrik.strobelt@ibm.com.
\item Jambay Kinley - Cornell University, New York, New York, United States.
\item Robert Krueger - Visual Computing Group, Harvard University, Cambridge, Massachusetts, United States.
\item Johanna Beyer - Visual Computing Group, Harvard University, Cambridge, Massachusetts, United States.
\item Hanspeter Pfister Visual Computing Group, Harvard University, Cambridge, Massachusetts, United States.
\item Alexander M. Rush - Cornell University, New York, New York, United States.
}

\shortauthortitle{Strobelt \MakeLowercase{\textit{et al.}}: GenNI: Human-AI Collaboration for Data-Backed Text Generation}

\abstract{Table2Text systems generate textual output based on structured data utilizing machine learning. These systems are essential for fluent natural language interfaces in tools such as virtual assistants; however, left to generate freely these ML systems often produce misleading or unexpected outputs. GenNI (Generation Negotiation Interface) is an interactive visual system for high-level human-AI collaboration in producing descriptive text. The tool utilizes a deep learning model designed with explicit control states. These controls allow users to globally constrain model generations, without sacrificing the representation power of the deep learning models. The visual interface makes it possible for users to interact with AI systems following a Refine-Forecast paradigm to ensure that the generation system acts in a manner human users find suitable. We report multiple use cases on two experiments that improve over uncontrolled generation approaches, while at the same time providing fine-grained control. A demo and source code are available at \url{https://genni.vizhub.ai}.
} %


\teaser{
  \centering
   \includegraphics[width=\textwidth]{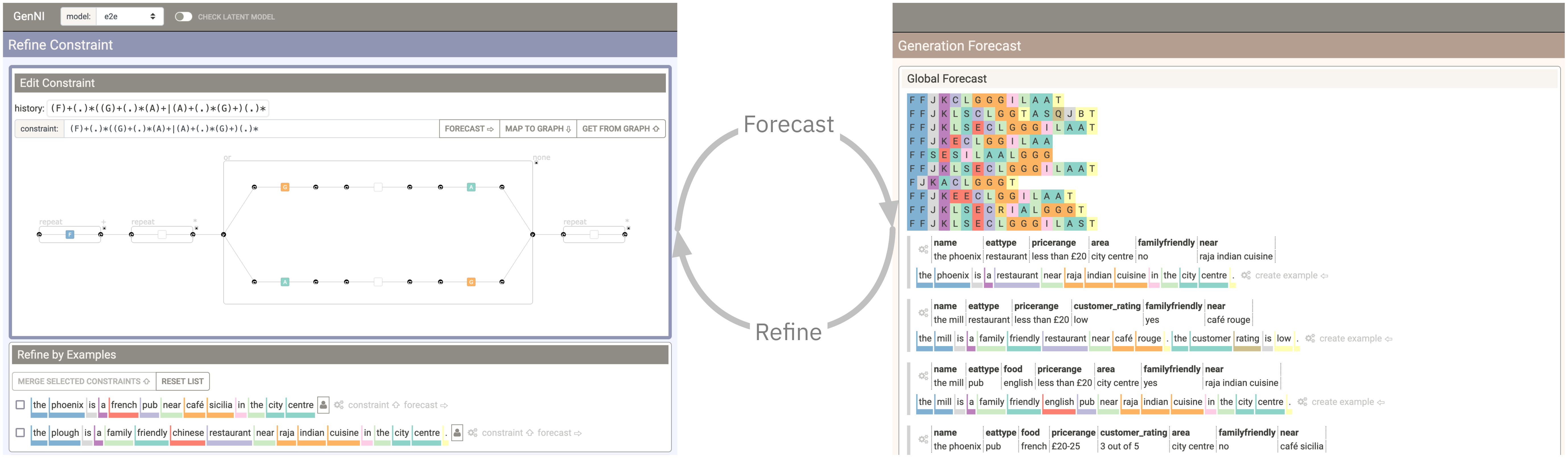}
  \caption{GenNI's forecast-refine loop applied to an example of a table2text model for generating
    restaurant descriptions of different forms. The right side shows the Generation Forecast Component with examples of unconstrained generation. The left side shows the Constraint Refinement Component with the construction of a potential new constraint graph.}
  \label{fig:teaser}
}

\vgtcinsertpkg

\begin{document}

\firstsection{Introduction}

\maketitle

Artificial intelligence methods for text generation are becoming increasingly advanced, with systems demonstrating convincing output in many surprising domains such as news and story generation. Machine learning based systems can learn how to generate text from seeing a massive amount of examples of human writing and interaction in the wild~\cite{radford_language_2019}. Underlying the systems are large models that process textual input and learn how to mimic the word-use, syntax, and high-level knowledge for generic generation. These tools can also be adapted to environments where they are tasked with generating specific conditioned textual outputs. For instance, in this work, we consider the problem of generated textual description of structured data, i.e., the Table2Text setting.

However, the ability to generate fluent output is often not sufficient for real-use cases. Many settings that use textual generation, such as user assistants (e.g., tools like Alexa) or automated response and search, require systems that can generate specific responses in a high-precision manner. Generic free generation systems cannot directly be deployed in these scenarios since they often ``go off-script'' and generate information that is not supported by their conditioning~\cite{wiseman_challenges_2017}. It is a non-trivial challenge to ensure that these systems stick to a specific type of conditioned generation, even when a  user clearly knows the goals and targets of the system. Given this issue, many approaches use human-crafted rule-based systems as opposed to machine learning.

GenNI is a prototype and a framework that facilitates Human-AI
collaboration for the challenging domain of data-backed text
generation with machine learning systems. The goal is to achieve the
benefits of ML based generation, while ensuring the precision of
human crafted control by using visual means. GenNI supports three
targeted aspects of the collaboration process:

\begin{itemize}
\item \textbf{Refinement of Model Constraints} GenNI allows users to impose model constraints that ensure specific high-level properties hold during generation.
\item \textbf{Forecasting of Model Outputs} GenNI makes it easy to see how these constraints affect real generation examples across a representative range of inputs.
\item \textbf{Deployable Model Corrections} The GenNI prototype utilizes a rigorous constraint-graph method that makes it easy to view the model update and utilize it in production use cases.
\end{itemize}

The system design is based on the collaborative framework of Gehrmann et al.~\cite{gehrmann_visual_2020}, who argue that it is critical to design ML systems that take the end-user user control scenario into account during the model design process. GenNI incorporates a controlled text generation model trained to interact with human users through explicit \textit{control states}. The tool facilitates an interactive and visual negotiation where a human user \textit{refines} the set of possible generations through a \textit{constraint graph} and then explores the system's actual outputs through a global \textit{forecasting} procedure. The tool allows the user to cycle through concrete examples to build up the constraint graph as they go.

In \autoref{sec:model} we introduce controllable text generation formally, describing the model and modes of user interaction. We then provide in \autoref{section:motivating-case-study} a guiding example for how a user might use the interactive tool for table2text description. \autoref{sec:goals-tasks-usergroups} introduces the goals and tasks required for building the prototype called GenNI for data-backed generation through Human-AI collaboration. \autoref{sec:design} presents the design decisions made to follow these guidelines and \autoref{sec:implementation} describes the implementation details of the tool. Further use cases for GenNI are given in \autoref{sec:case-study}.
\autoref{sec:related-work} reviews the related work in this research domain to provide some context for our work.
We conclude the paper in \autoref{sec:conclusion} by outlining ideas for future directions.

\section{Model: Table2Text with Controls}
\label{sec:model}

Underlying GenNI is a model designed for controllable text generation to enable visual interaction. The model extends standard ML models for text generation with explicit \textit{control states} (i.e., discrete latent variables) that allow an end-user to alter the model's output through constraints. Additionally, given an input and output, the model provides a method for inferring the control states. These control states are the main interface used by GenNI.

Table2Text generation aims to produce a textual \textit{description} consisting of word tokens $\vy_{1:T}$ from an input of data $\vx$  represented as a table. While one could generate this description directly from the data, we distinguish controllable systems as ones that provide intermediate control states for directing the structure of the description.  We use one control state for each word, $\vz_{1:T}$, that are generic discrete values from a small label set, for example, represented as letters A to Z. Each control state corresponds to a high-level cluster of the corresponding word's semantics learned by the model for the problem. An underlying assumption of this work is that an end-user can craft higher-level constraints with these states than by acting directly on words. 

\begin{figure}[h!]
    \centering
    \includegraphics[width=.9\linewidth]{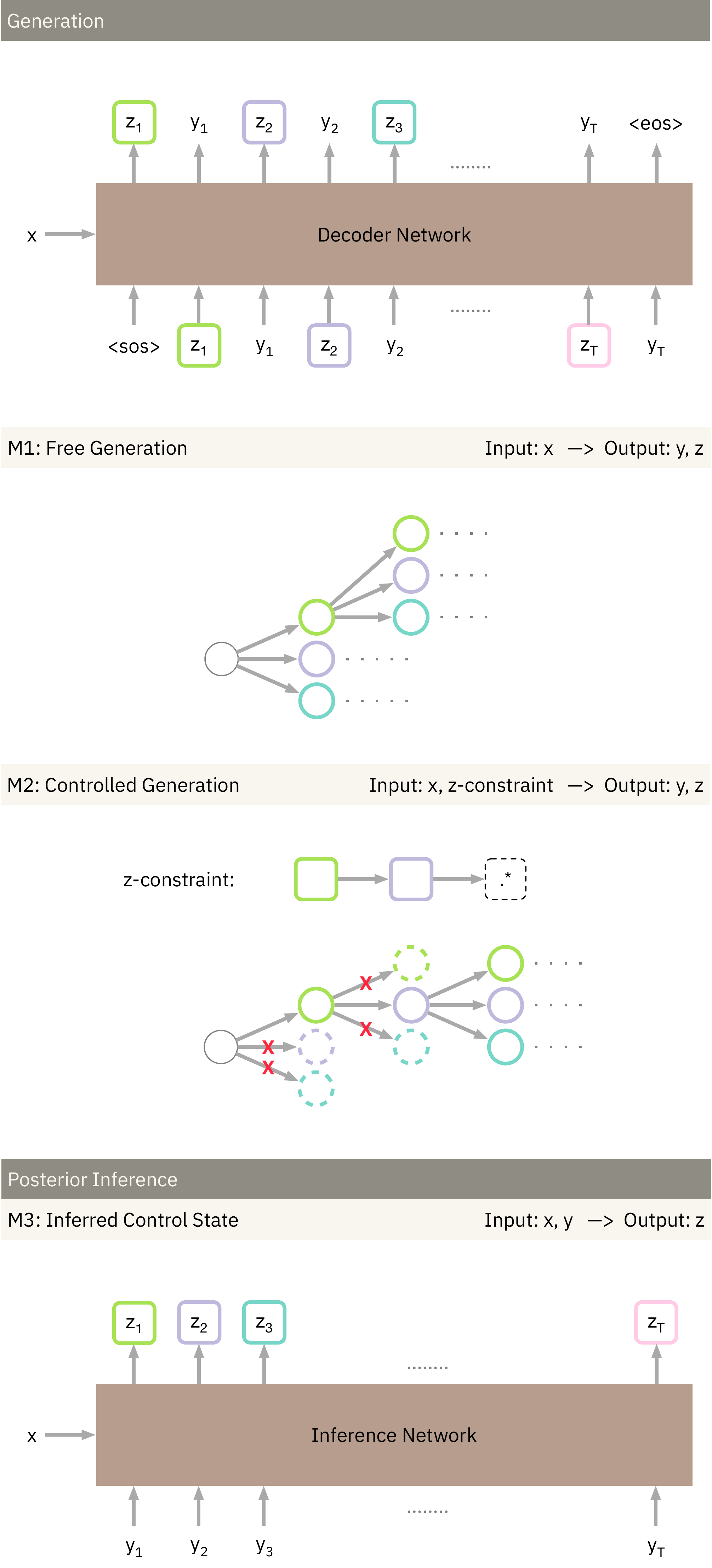}
    \caption{The different modes (M1-M3) of the model which are the building blocks that enable visual interaction for GenNI. Modes M1 and M2 are  outputs $\vy$ and $\vz$ inferred from $\vx$ by free and controlled generation. Mode M3 describes the inference of control states $\vz$ from data $\vx$ and text $\vy$.  (see \autoref{sec:modelinteract})}
    \label{fig:model}
\end{figure}

Formally, the model outputs control states and one word at a time. Starting from data $\vx$, the model generates the description $\vy$ auto-regressively (left-to-right) by first generating a state $\vz$ and then the corresponding word.
\begin{align*}
\vz_1 &\sim p(\vz_1\ |\ \vx)  &    \vy_1&\sim p(\vz_1\ |\ \vx, \vz_1) \\
  \vz_2 &\sim p(\vz_2\ |\ \vx, \vz_1, \vy_1) &  \vy_2&\sim p(\vy_2\ |\ \vx, \vz_1, \vy_1, \vz_2)\\
                                                       \vdots
\end{align*}
This process produces the probability of the description and states given the data $p(\vy, \vz | \vx)$. Each part is implemented using a deep learning model. We utilize a recurrent neural network to predict both the $\vz$ and $\vy$ outputs and an attention based encoder to condition on the input $\vx$. Specifically the probabilities are given as:
\begin{align}
p(\vz_t | \vz_{<t}, \vy_{<t}) &= \text{softmax} (\mW_0 \vh(\vx, \vy_{<t}, \vz_{<t})) \\
p(\vy_t | \vz_{\leq t}, \vy_{<t}) &= \text{softmax} (\mW_1 [\vh(\vx, \vy_{<t}, \vz_{<t}) + \vg(\vz_t)])
\end{align}

\noindent where $\vh$ is the output of a recurrent neural network over the input, previous words, and control states, $\mW$ are parameters, and $\vg$ is a function of the current control state.

A key aspect of the system will be the reverse process, i.e. inferring control states $\vz$ from examples tables and sentences $\vx, \vy$ . 
 Unfortunately, the posterior distribution $ p(\vz | \vx, \vy)$ is intractable to compute exactly. We therefore employ variational inference to approximate this distribution using a parameterized \textit{inference network} $q(\vz | \vx, \vy)$. We train this approximation jointly with the forward model in the standard variational autoencoder framework~\cite{kingma_auto-encoding_2014, mnih_neural_2014, rezende_stochastic_2014}.

For the inference network, we use a linear-chain conditional random field (CRF) with a neural parameterization. This family of distributions is particularly suited for labeling segments of the text with control states,
and has been used effectively in similar tasks~\cite{li_posterior_2020, fu_latent_2020}. To train this part of the system, we use the Gumbel-CRF method proposed by Fu et al.\cite{fu_latent_2020} which allows us both to learn the variational approximation and also train the rest of the model with concrete control states.

Finally, during training we enforce a soft correspondence between control states and table properties. These constraints enforce weak supervision in the form of a heuristic alignment between data and text, i.e. whether some part of the sentence is describing a table field using identical text as in the table. Following Li et al. \cite{li_posterior_2020}, we use a technique known as posterior regularization which allows the model to follow or ignore these alignments. We find this can produce more human-legible control states for some of our tasks.

\subsection{Model Interaction}
\label{sec:modelinteract}

Directly visualizing deep learning models is challenging~\cite{hohman_visual_2019}.
Instead of visualizing the internals of the model, we interact
with it through its outputs and control states through three distinct modes shown in \autoref{fig:model}: (M1) \textit{free generation} where the model searches for the most likely textual output given the input data; (M2) \textit{controlled generation} where the model searches for an output that obeys a constraint graph; (M3) \textit{inferred control} a reversed version of generation, where the user provides a goal text, and the control states are inferred. These three modes will act as the building blocks for the interactive system.

\noindent
\textbf{Free Generation (M1)}: The most basic operation is to allow the AI model to generate freely. This mode produces the highest-scoring output from the model, formally $\arg\max_{\vy, \vz} p(\vy, \vz | \vx)$. Note, though, that high scores from the model often do not correspond to generated text that a human user would have wanted.

Practically, computing this argmax is a search problem over a very large search space. It is common to utilize an algorithm known as \textit{beam search} as an approximation. Beam search works by exploring the search tree using a fixed number of hypotheses per time step, i.e., considering five different hypothesis sentences of $t$ words long before moving on to five different hypothesis sentences of $t+1$ words. We can extend the beam search for controlled generation to alternate between exploring the best succeeding control state and the best next word. This search is shown in \autoref{fig:model}~(top).

\noindent
\textbf{Controlled Generation (M2)}: An alternative to free generation of the output text is to control the generation to fit specific cases of an end-user application. For this mode, the user provides an explicit constraint graph to the model. These constraints are applied to the control states $\vz$, which act to restrict the word outputs $\vy$. For controlled generation, we solve $\arg\max_{\vy, \vz\in {\altmathcal Z}} p(\vy, \vz | \vx)$ where ${\altmathcal Z}$ is the constrained set of possible outputs.

Formally, the constraint graph is equivalent to a regular expression on control states restricting the set ${\altmathcal Z}$. Regular expressions allow a user to encode complex sets for shaping the space of possible outputs. To leverage this control, we use a constrained beam search algorithm where the constraint graph ensures that $\vz$ is correct for the model output (\autoref{fig:model}~(middle)).
For instance, given the constraint \texttt{A.B*}, during generation, we ensure that the control state of the first word is \texttt{A}, the second is unconstrained \texttt{.}, and the remaining words must have control state \texttt{B}.

It is important to note that these constraints do not directly constrain the words generated $\vy$. Each control state can generate many possible words. This can be seen in \autoref{fig:design} (g) where, given a control state, the model has options over the word to generate. 

\noindent \textbf{Inferred Control (M3)}: As described above, the system is trained to also allow for reverse computation of control states, using an inference network. This mode allows for the inferred control where the user can write an expected output sentence, and the model will produce plausible control states $\argmax_{\vz} p(\vz \ | \ \vx, \vy)$. While generation allows us to infer output given an input, this term allows us to find the control states for a human written input and output (\autoref{fig:model}~(bottom)).

\begin{figure}[ht]
    \centering
    \includegraphics[width=\linewidth]{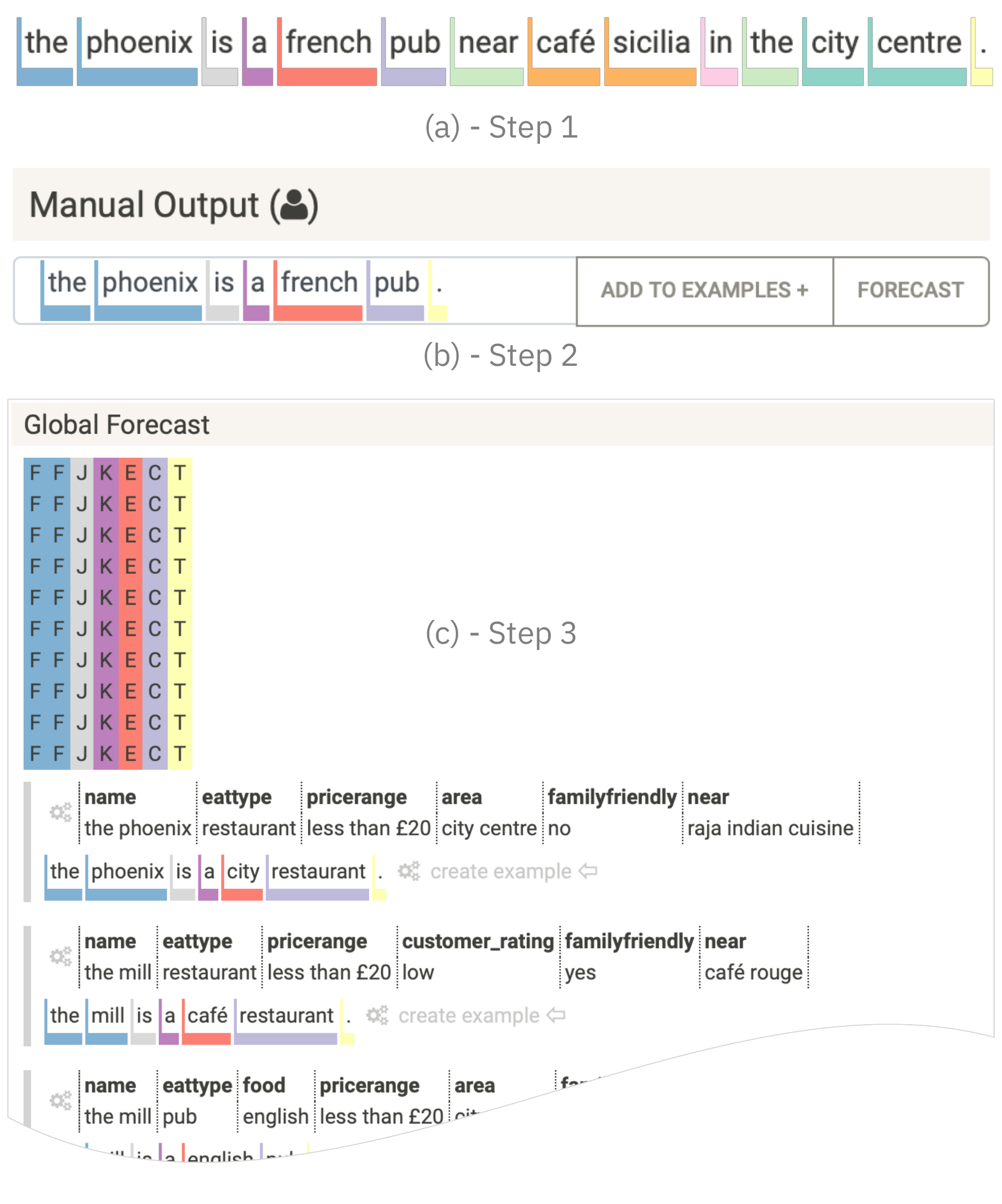}
    \caption{Motivating use case, steps 1--3. (a) Model output during free generation (M1). The control states are indicated by color below the produced output tokens. (b) Alice provides the custom output ``\texttt{the phoenix is a french pub .}'' The matching control states are inferred and mapped to colors (M3). (c) The constraint ``FFJKECT'' is applied to other random inputs (M2).}
    \label{fig:iuc1_3}
\end{figure} 

\section{Motivating Case Study: Collaborative Generation}
\label{section:motivating-case-study}

We now consider a motivating case study of human-AI collaboration for text generation in GenNI. Alice is a user building an ML-based chatbot system. She is designing a module that generates a restaurant description from a table~\cite{novikova_e2e_2017} such as the following:

\vspace{2pt}
\begin{tabular}{c | c | c |c | c}
     \textbf{name} & \textbf{eat type} & \textbf{food}  & \textbf{area} & \textbf{near}  \\
     the phoenix & pub & french & city center & cafe sicilia
\end{tabular}
\vspace{2pt}

Alice first tried developing a system using free generation (M1) but found that the generated text is highly variable and does not suit her specific use case, which requires high precision constraints. In particular, Alice wants the system to focus exclusively on the cuisine (food) and the type of establishment (eat type). However, she found that, under free generation, the model would use all of the fields, e.g., generating:

\vspace{2pt}
\noindent\texttt{the phoenix is a french pub near cafe sicilia in the city centre .}
\vspace{2pt}

To benefit from our controllable system, Alice needs to provide specific constraints on the control states to produce high-precision outputs that fulfill her specified goals. Concretely, she will want to develop a constraint graph that ensures that her system outputs are correct. We describe her use of GenNI to achieve this goal.

\noindent \textbf{Step 1: Observing Control States on an Example}
To begin to gain intuition into model prediction and the control states, Alice starts with a specific example input $\vx$, e.g., the table row shown above. GenNI produces text in free model generation (M1) while also showing the control states for each word token -- \autoref{fig:iuc1_3}a.

This step grounds the collaboration process in a specific starting point. This visual representation maps concrete textual outputs to the underlying control states. For this example, Alice infers that the model has allocated the blue state for the restaurant name, red for food type, and cyan for location.

\noindent  \textbf{Step 2: Inferring Control States for Manual Output} Next, Alice can actively posit a counterfactual: ``What would the control states have been for a user-generated target output?''  To do this, Alice can provide her own textual description (``\texttt{the phoenix is a french pub .}'') to the system in the form of the sequence $\vy_{1:T}$. Utilizing the model's inference mode (M3), the model will infer the control states that would have most likely lead to that output -- \autoref{fig:iuc1_3}b.

This step is the start of the \textit{refinement} procedure that allows Alice to build up a constraint graph on the model itself. Specifically, if she is happy with the control states that the model assigned, she can add this sequence of control states to the constraint graph. This refinement tells GenNI that this sequence of control states is appropriate for the AI model to generate.

\noindent \textbf{Step 3: Forecasting AI Generations Under Constraints} The constraint graph allows Alice to ensure that the model generates outputs from a set of acceptable control state sequences. However, it can be difficult to tell how constraints will generalize across examples. GenNI allows users to do this through \textit{forecasting}, applying controlled generation (M2) across a set of different diverse inputs. In this case, GenNI allows Alice to randomly sample different inputs with different properties to observe generations from the system or, alternatively, probe a range of targeted inputs -- \autoref{fig:iuc1_3}c.

Alice can then view all these outputs simultaneously to observe patterns and relationships. She confirms her hypothesis about the control states for the restaurant name (blue) and food type (red). The \textit{forecasting} feature of GenNI makes it easy for her to see specific regions where the AI failed to generate the correct output.

\noindent  \textbf{Step 4: Precision Refinement of Constraints}: The set of control states obtained in Step 2 can be applied as a constraint for other inputs (see Step 3).  However, this constraint is specific to the input used to obtain it and may not generalize well. Alice can use a regular expression editor to refine the constraint so that it may generalize better.

\begin{figure}[t]
    \centering
    \includegraphics[width=\linewidth]{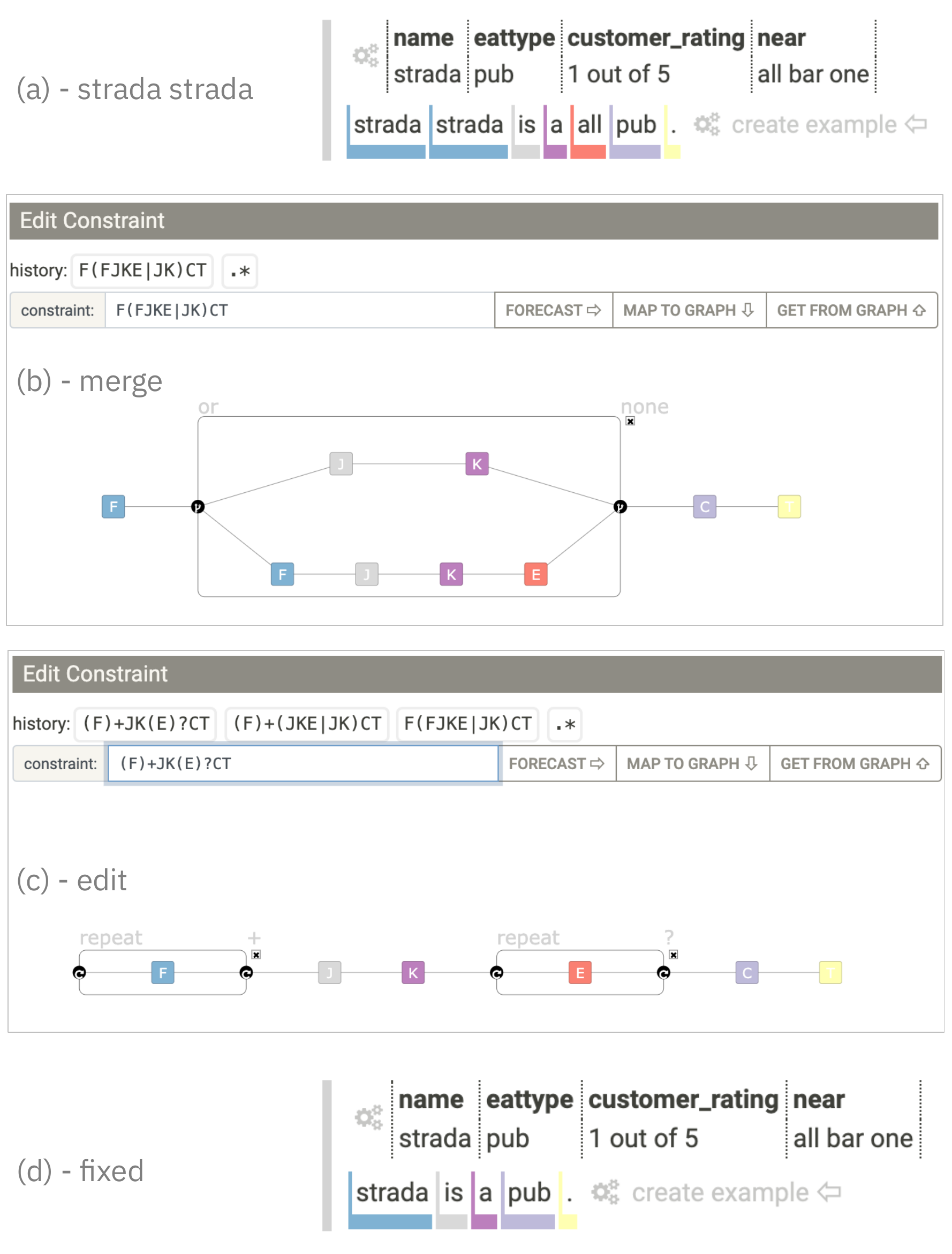}
    \caption{Motivating use case, step 4. (a) Alice observes that ``strada'' is copied twice and no information about cuisine is available. (b) The selected example sequences of control states are merged into a combined constraint graph shown in the Constraint Editor view. (c) The constraint graph can be edited in the Constraint Editor by either using the text editor or the visual editing tool. (d) After applying the refined constraint, Alice observes the correct output: ``strada'' is copied once and no food type is mentioned.}
    \label{fig:iuc4}
\end{figure}

When browsing through the forecast results, Alice spots a particular problem. The model copies the establishment name ``strada'' twice from the input and also fantasizes a food type -- \autoref{fig:iuc4}a.

Alice knows from her exploration that the blue control state instructs the model to copy the establishment name. However, her current constraint is too rigid, since some restaurants like ``strada'' are only a single word (vs. ``the phoenix''). Similarly, some tables do not reveal the food type. By switching to this example and correcting (using step 2) for the repetition, Alice can include the corrected sequence of control states into the constraint graph to allow for both outputs -- \autoref{fig:iuc4}b.

Alternatively, using the constraint editor, Alice can manually replace the sequence of two blue boxes to a variable length repeat of blue boxes and make the red box optional -- \autoref{fig:iuc4}c.

After applying the newly created constraint, she can forecast again and confirm that the issue has been fixed -- \autoref{fig:iuc4}d.

\noindent  \textbf{Step 5: Building Constraints into a Model}:
Alice can repeat this process of forecasting and refinement to obtain a constraint that generalizes well. She repeats Steps 1-4, each time alternating between observing generation, inferring control states for custom outputs, transferring across input types, and merging constraints. In each iteration, the constraint graph grows with the addition of more rules. Finally, Alice can save the constraints to use the constrained model on a more extensive test set and in production.

\begin{figure*}[h]
    \centering
    \includegraphics[width=.8\textwidth]{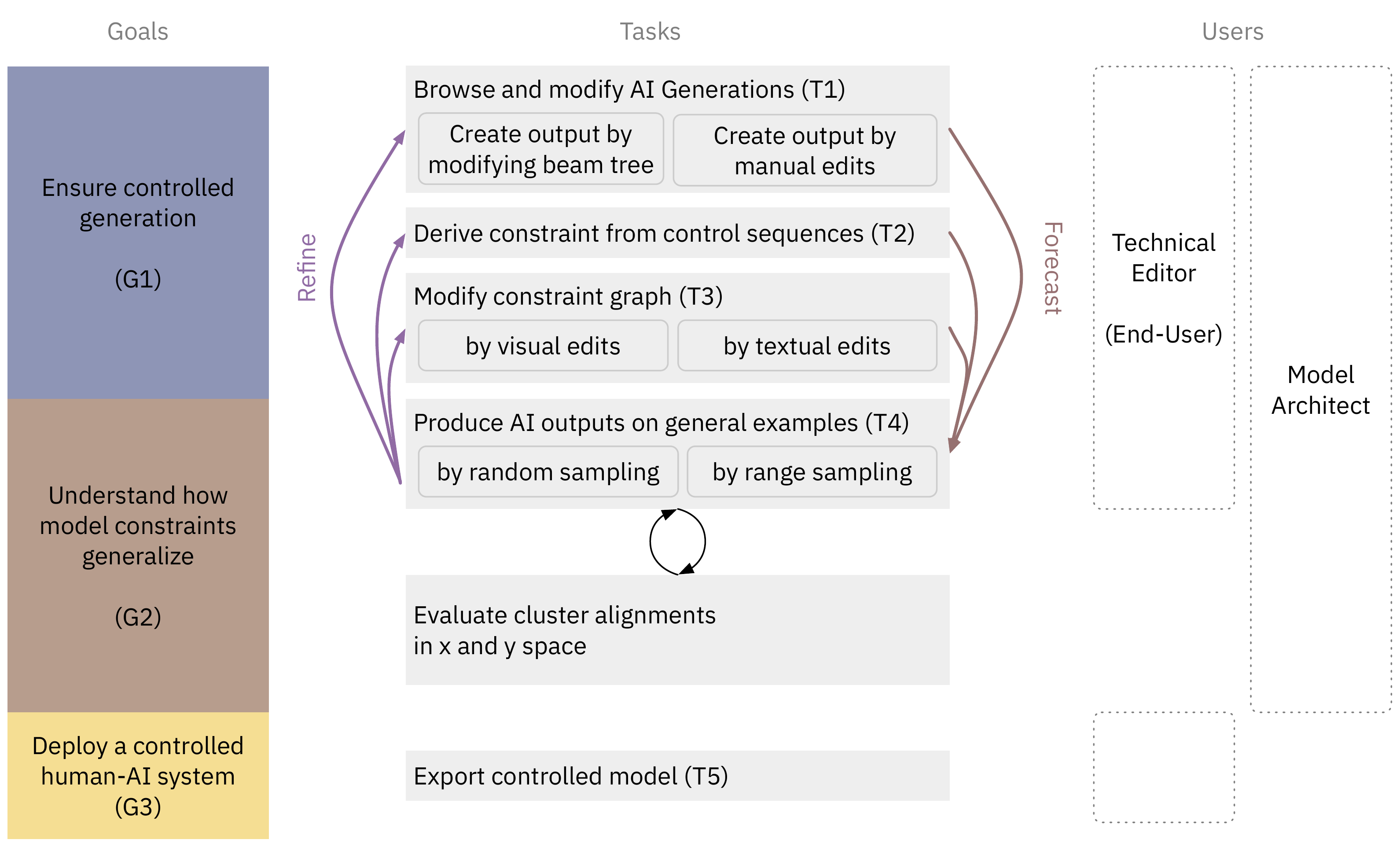}
    \caption{Overview of GenNI domain goals, interaction tasks, and addressed user groups. GenNI aims at supporting goals for working with control state models: (G1) ensuring controlled generation, (G2) evaluating these constraints on subsets of relevant data and demonstrate model constraint generalization, and (G3) deploying the outcome as a controlled Human-AI system. From these high-level goals, a series of interaction tasks (middle part) is inferred. These tasks are the building blocks for the main interaction loop of forecasting constraint effects and refining constraints as a result. GenNI targets end-users and model architects alike for most tasks (right). (see \autoref{sec:goals-tasks-usergroups})}
    \label{fig:goals}
\end{figure*}

\section{Goals, Tasks and User Groups}
\label{sec:goals-tasks-usergroups}

GenNI aims to support the collaborative development of data-backed generation systems. AI tools for generation can efficiently produce textual outputs on a variety of inputs; however, without close inspection of these free generations, it is difficult for a human user to find issues or correct errors. Alternatively, human users can produce careful example outputs, but each is slowly crafted and hard to generalize.

At each alternating round of GenNI's use, either the user can formulate explicit constraints on the AI system, or the AI system can generate a set of outputs based on its current state. In this manner, the user can quickly observe that the system is over-constrained or under-constrained while at the same time having assistance from the AI system to help produce generalizable constraints. When the collaboration is over and both sides reach an equilibrium, the full set of constraints produced in the process can be incorporated into the AI system.

To act as a tool for reaching this human-AI equilibrium for constraints, GenNI was designed with three high-level user goals in mind (\autoref{fig:goals}):

\noindent \textbf{G1: Ensure controlled generation} AI systems with free generation can produce unexpected outputs which do not follow the guidelines that the user prefers. The goal is to provide feedback controls in the form of constraints. The language of control states may be difficult for a user to apply directly, so the tool must convey how these work and make it easy for the user to link these to specific examples. Once constraints are created, a user needs to be able to manipulate the constraints in an intuitive and precise manner.

\noindent \textbf{G2: Demonstrate model constraint generalization}
Upon specifying constraints, the user needs to understand how the AI will apply and interpret them in a global manner. While the user may have an intuitive sense of the constraints, they will not know whether they will act consistently and naturally across any input the AI receives. As such, a tool needs to provide guidance about general outputs such that a user can build intuition and trust the system.

\noindent \textbf{G3: Deploy as a controlled Human-AI system} Many approaches for debugging neural models find issues but do not provide a path for remediation. The final goal of GenNI is to produce a constraint set that can be packaged and deployed as part of a production model. The constructed constraint graph contains all of the final information about the appropriate controlled use of the system, and controlled generation can be efficiently run on real systems. After deployment, if an issue comes up in production use,  new constraints can be refined into the model.

The case study in \autoref{section:motivating-case-study} takes advantage of these goals. The user is interested in the targeted use of the generation system in Step 1. Step 2 is a step towards defining constraints based on the user goals (G1). These constraints need to be explored on a larger set of examples (G2) in Step 3. However, upon observation, further refinement is needed in step 4 (return to G1). Finally, the negotiation leads to a model in Step 5 that can be deployed in a production system (G3).

These three domain goals motivate the main interaction and tasks of the system:

\noindent \textbf{Task 1: Browse and Modify AI Generations} by observing the textual output $\vy$ and control states $\vz$ for possible inputs $\vx$. The user should be able to modify outputs manually or by using alternative model predictions. The user should develop an intuition about the control states and about the variety of outputs the model generates [G1, G2].

\noindent \textbf{Task 2: Derive Constraints from Control Sequences} in order to produce controls on the AI system. The user should be able to define an initial constraint graph from various preferred examples [G1]

\noindent \textbf{Task 3: Modify Constraint Graph} to allow for finer grained control of the final constraints. Some generalizations cannot be derived from examples directly but require user adjustments. These should be given enabled via a textual and a visual interface. [G1]

\noindent \textbf{Task 4: Produce AI Outputs on General Examples} to observe the generalization behavior and to confirm the correctness of the systems. A user should be able to forecast what the current change of model constraints would mean on a more global scale. This should be at close to interactive rates. [G2]

\noindent \textbf{Task 5: Export controlled model} to allow deployment of the model in production or for broader testing. [G3]

GenNI targets two user groups (see~\cite{strobelt_lstmvis_2018}). One group is a technically versed end-user that does not need to know about the underlying model, just about the task at hand. We call this group \textit{technical editors}. The other group is \textit{model architects} that want to evaluate their model under real human constraints. See \autoref{fig:goals} for reference.

\begin{figure*}[h!]
    \centering
    \includegraphics[width=.9\linewidth]{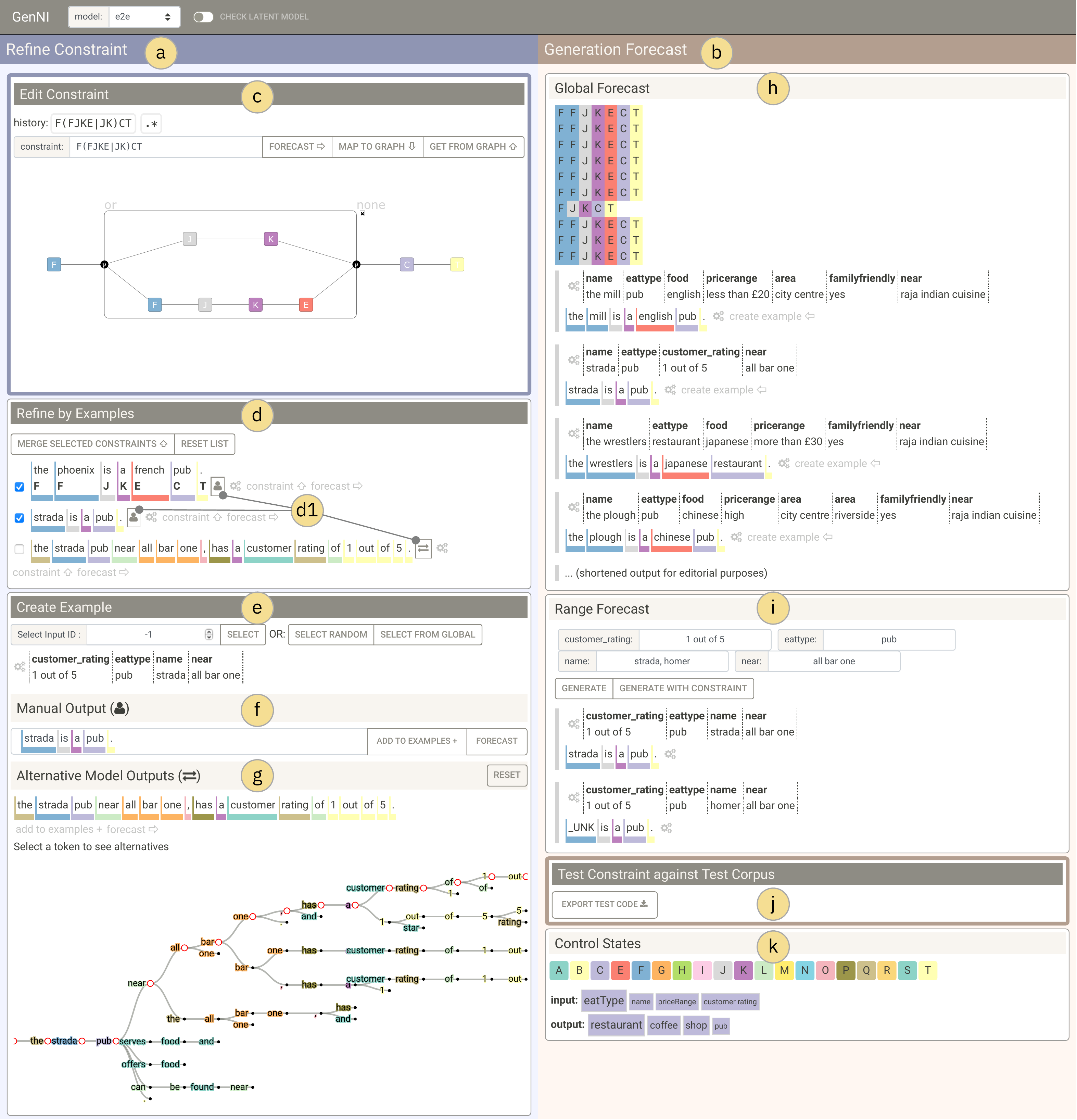}
    \caption{The GenNI user interface is split into a Constraint Refinement component~(a) and a Generation Forecast component~(b). A user can edit constraints directly in the Constraint editor~(c) or derive them from a set of examples~(d). Examples can be created~(e) by inferring control states from a user output~(f) or from the model's beam tree~(g). The effect of specific constraints can be forecast on random samples~(h) or tested on data ranges~(i). The controlled model code can be exported for deployment or further testing~(j). Model architects can investigate the control state alignments for each control state~(k). Details are given in \autoref{sec:design}. }
    \label{fig:design}
\end{figure*}

\section{Design}
\label{sec:design}

GenNI is an interactive prototype for facilitating collaborative interaction for controlled generation. The visual layout and the interactions are the results of an iterative design process between visualization experts, NLP researchers developing controllable models, and users aiming to deploy these models in practice.

The GenNI interface is constructed out of two meta components that immediately reflect the domain goals G1 and G2. These components are juxtaposed to facilitate the continuous iteration between forecasting constraints on global examples and refining the constraint as a result of observing global effects. Accordingly, the left half collects Refine Constraint views. (\autoref{sec:refineview}) The right half provides Generation Forecast views. (\autoref{sec:forecastview})

\subsection{Constraint Refinement Component}
\label{sec:refineview}

The Constraint Refinement component (\autoref{fig:design}a) facilitates the construction and exploration of constraints by the user (G1). It allows direct editing of constraints and constructing constraints from well-crafted examples. This component has three supporting views: 1) a Constraint Editor (\ref{fig:design}c) which allows direct textual and visual modification of constraints, 2) a Refine-by-Example (\ref{fig:design}d) view to collect and utilize examples for constraint refinement, 3a) an Example Creation view (\ref{fig:design}e) to construct output examples by manual edits (\ref{fig:design}f), and 3b) to generate examples utilizing alternative model decisions (\ref{fig:design}g).

The \textit{Constraint Editor} (\autoref{fig:design}c) encodes the entire collaborative state of the system, i.e., all information collected to constrain the generation procedure. A user can add multiple different valid control state sequences, and they will be merged into this graph [T2]. The constraint graph is represented in two ways. First, it is encoded as an editable node-link diagram.  The user can modify and update the constraint by visually adding, deleting, and modifying nodes and meta-nodes (like OR) [T3]. Secondly, the constraint graph is represented by a  simple language borrowing the syntax from regular expressions. At the top of the Constraint Editor, this textual representation can be modified directly [T3]. For keeping track of provenance, a history field collects previous iterations of constraint formulations in textual form. On click, they are available for re-editing.

To develop new constraints from user preferred examples, the user needs a place to collect examples and merge them into complex constraints. The \textit{Refine-by-Examples} view (\autoref{fig:design}d) provides these functions. Each example represents $\vy$ as text and $\vz$ as a color. Their origin (i.e., human-generated or model alternative) is encoded as a postfix symbol (\autoref{fig:design}d1) followed by buttons to trigger propagation to the Constraint Editor [T2] or forecasting in the Global Forecast view [T4] (see \autoref{fig:design}h).
If multiple examples are selected, the Merge Constraints button attempts to merge all of them into one combined constraint graph for the Constraint Editor [T2].

Examples are created by using the Example Creation view [T1] (\autoref{fig:design}e).  The user starts by first setting a reference input $\vx$. This input can be acquired by selecting an input ID to point to one item in the model's test set directly or randomly. Alternatively, examples can be selected from the Global Forecast view.

To produce a matching output, the user can write a custom freehand text $\vy$ (Manual Output, \autoref{fig:design}f) and derive the matching control states $\vz$ from the inference network (M3) of the model.

A second way to produce outputs is by meaningfully interacting with the model internals to modify its predictions (\ref{fig:design}g). The user can create these alternative model outputs by constraining the beam tree (lower part) from the visual tree representation or by selecting alternative tokens by clicking on a token. The beam tree tool allows the user to see the paths taken by the model during beam search, probe its decisions at specific locations and even alter its decisions (when choosing the control state) to see in real-time the effect of changes to the constraint graph.  While this view might be complex for the end-user group, it provides a way to generate outputs that the model can reproduce.

In both cases, the control states sequence of the produced outputs defines a simple constraint and can be tested by forecasting it, or it can be added to the Refine-by-Example collection to create a new constraint graph.

\subsection{Generation Forecast Component}
\label{sec:forecastview}
The right-hand component of the GenNI (~\autoref{fig:design}b) visualizes the model's response to the constraints. It presents a global insight into the AI system by providing either free or controlled generation on a more extensive set of examples [T4].

The Global Forecast view (~\autoref{fig:design}h) conveys a global perspective on the effect of constraints and utilizes random sampling from the test set to produce different $\vx$ values, which are input to constrained generation and produce $\vy$ and $\vz$ values. This sampling and generation results are shown as a tuple of two rows containing input table and output text with color highlighting. Each tuple can become the next reference example for the Example Creation view and, in this way, contribute to the refinement of the constraint. All $\vz$ values are summarized in a heatmap on top of the view to better see an alignment between constraint outputs.

The Range Forecast view (~\autoref{fig:design}i) provides the same features like the Global Forecast view, but the $\vx$ values are selected from value ranges or lists of values. This allows more systematic testing in a local neighborhood of examples. E.g., for the use case of producing date strings (see \autoref{sec:case-study}), a user could test which influence the day value has by producing  $\vx$ and generations for all days of a specific month.

\subsection{Encodings}

The concept of a control state is central to the functioning of the collaboration and provides a shared space between the human user and the underlying AI system. All constraints are developed on these states, and they are the single unit of transfer between the two views (see ~\autoref{fig:design}). As such, GenNI uses a visual encoding of control states as colors in all locations, both in the constraint formulation and the evaluation side. Unlike words, which are very fine-grained, control states allow for a high-level color encoding. This visual encoding makes it easier for a user to see differences and anomalies in sequences quickly.

The use of color as a central encoding poses some design challenges. Legibility is drastically decreased if the contrast between background and text color is low. That restricts the use of colors to either very dark or very light palettes. To be less restricted in color choices, we changed our encoding for the combination of $\vy$ and $\vz$ from full background coloring to only color underlines. Only in the very space-limited beam tree view, we use color bleeding.

\begin{figure}[t]
    \centering
    \includegraphics[width=.8\linewidth]{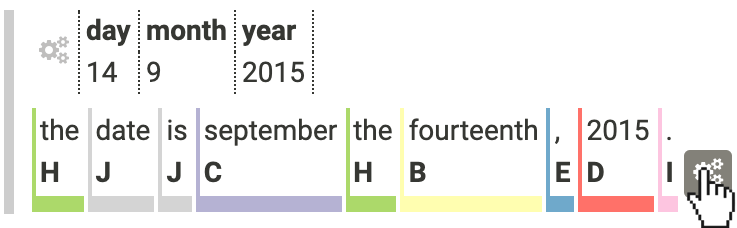}
    \caption{On click, the control states are represented by letters to support formulation of constraint graphs and to support color-blind users. }
    \label{fig:cluster_letters}
\end{figure}

Since color encoding is a core part of our prototype, we also thought about methods to support two scenarios where the color encoding might not be sufficient: 1) when modifying the constraint graph by textual input, the user has to refer to the colors in a meaningful way; 2) our color choices are not colorblind-safe. To address these issues, we added an optional representation of control states as letters. On click, the user can reveal the letters for single generation tuples. See \autoref{fig:cluster_letters}.

The selection and arrangement of all functional parts of the GenNI interface underwent many iterations. E.g., during the experimentation phase, all views were organized as rows to a single vertical list. The idea of juxtaposing Constraint Definition and Generation Forecast and arranging the subviews to support this bifold character results from understanding the interplay between forecast and refinement as a forth-and-back loop and not as a strict sequential order first-a-then-b.

\section{Use Cases}
\label{sec:case-study}

We apply the GenNI prototype to build controllable generation systems for two different domains. In both cases, a model architect utilized the system and explored the insights it gives for the problem and the underlying model. First, we build a model for a date conversion problem, where the model is simple enough such that all constraints can be explored. Next, we apply it to a real-world system using the E2E restaurant recommendation dataset~\cite{novikova_e2e_2017}.

\begin{figure}[tb]
    \centering
    \includegraphics[width=\linewidth]{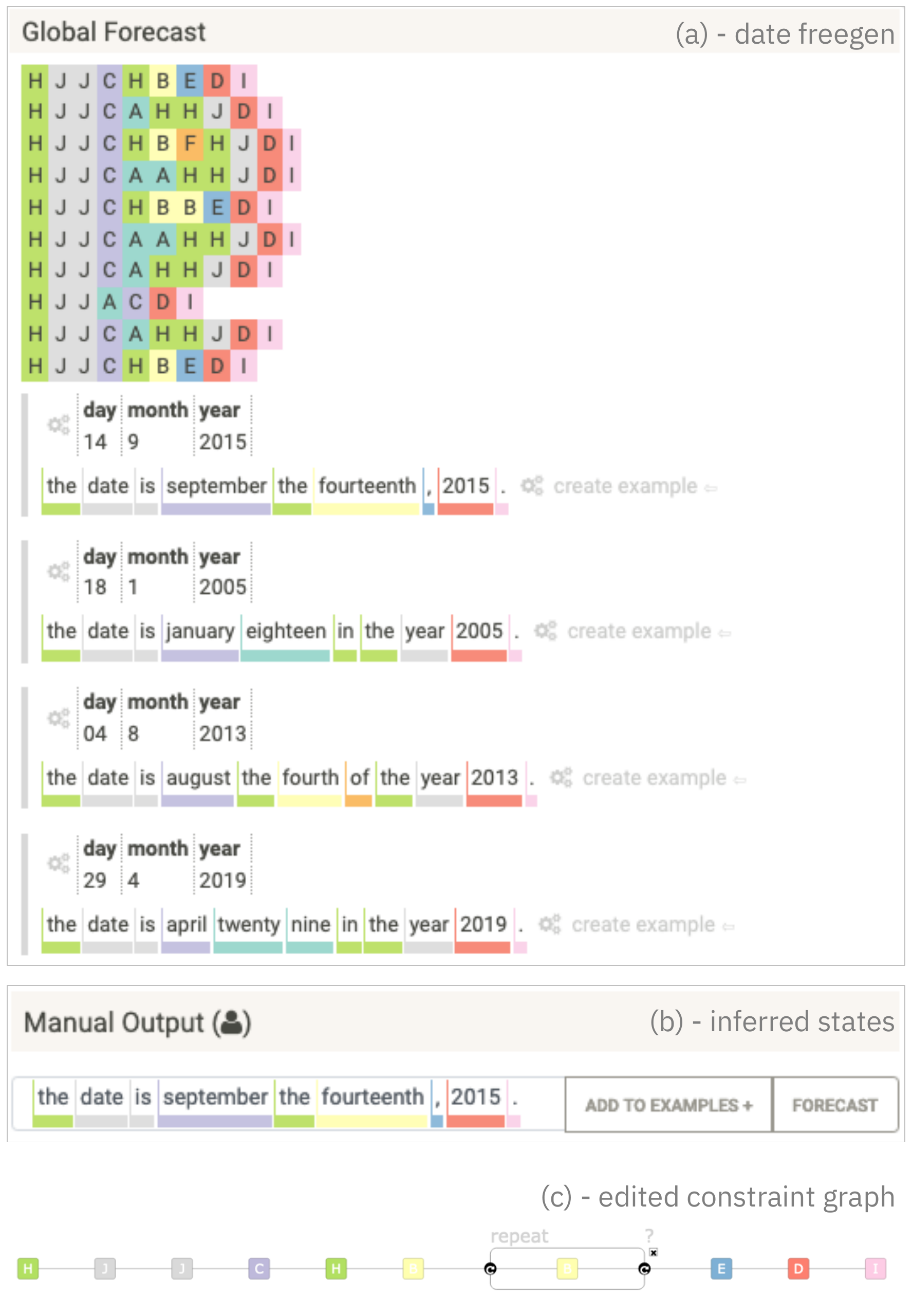}
    \caption{Date Generation use case. (a) Model outputs during free generation [T1]. (b) Inferred control states for provided text output [T2]. (c) Constraint graph inferred and edited for this output format [T3]. }
    \label{fig:date_uc_all}
\end{figure} 

\noindent \textbf{Date Generation}
The first model is a synthetic date generation dataset where the input $\vx$ is a table representing a date consisting of (day, month, year). The corresponding output text is a sentence $\vy$ describing the date. Eight different formats of representing the same date data were created using nominal or ordinal days, changing the order of day and month, and deciding to use commas or not before the year. For example, consider the input:
\vspace{2pt}
\begin{center}
\begin{tabular}{c | c | c }
     \textbf{day} & \textbf{month} & \textbf{year}\\
     14 & 9 & 2015
\end{tabular}
\end{center}
\vspace{2pt}

\noindent This date can be generated as ``\texttt{today is the fourteenth of september, 2015 .}'' or ``\texttt{today is september the fourteenth in the year 2015 .}'' as well as six other formats. Model control is used to select the preferred output form, e.g., the ordering of the days and months, use of commas, and ordinal vs. numerical ways of writing the day.

Under this well-specified task, the goal is to test if GenNI allows for reasonable clusters of words for the control states. We also want to see if we can construct constraint graphs for formats that generalize as expected. In particular, the text format allows for variable day lengths, so the constraint graph must allow for this output. GenNI provides tools for performing these tests through interactions with the AI system.

  \begin{figure*}[ht]
    \centering
    \includegraphics[width=\linewidth]{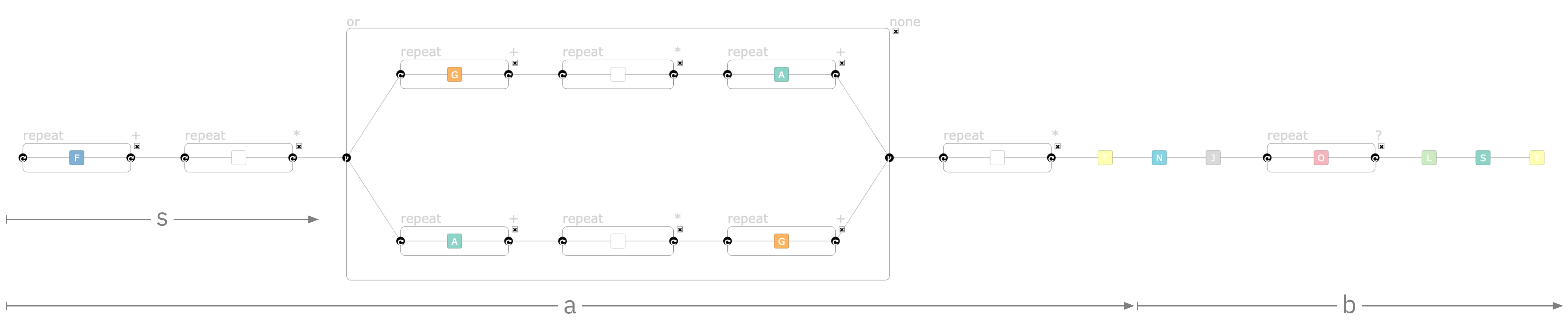}
    \caption{Final constraint graph produced in the Restaurant use case. Section (a) ensures generation of a location either in the order (\texttt{area}, \texttt{near}) or (\texttt{near}, \texttt{area}). Section (b) forces the use of \texttt{family-friendly} allowing an optional state for \texttt{not}. Section (s) is the seed that allows for additional descriptive information. }
    \label{fig:ex_location}
  \end{figure*} 
  
  \begin{figure}[ht]
    \centering
    \includegraphics[width=0.9\linewidth]{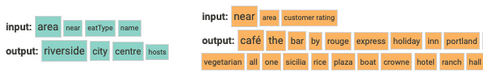}
    \caption{Alignment between control states, table field and text. Cyan is the control state for \texttt{area}. Orange for the \texttt{near}.}
    \label{fig:ex_align}
  \end{figure}
  
  \begin{figure}[ht]
    \centering
    \includegraphics[width=0.9\linewidth]{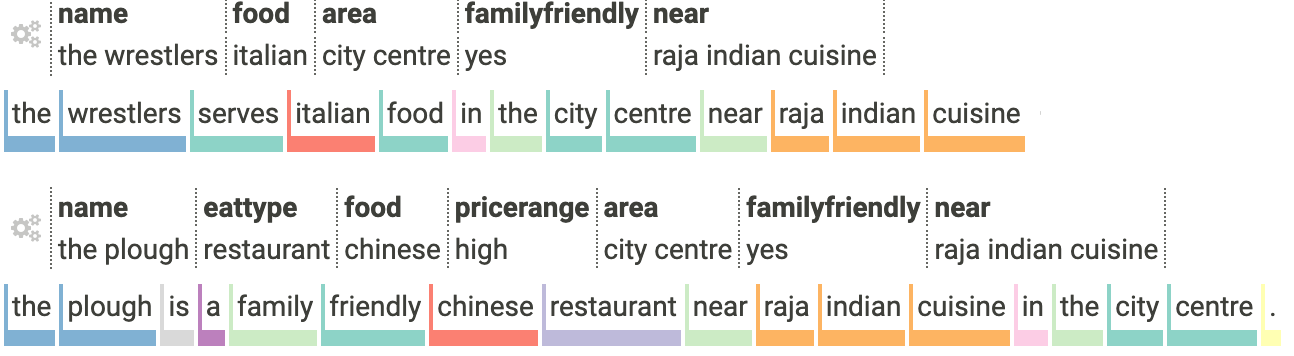}
    \caption{Generation output for constraint graph section (a) in \autoref{fig:ex_location}. Note that the 
        first output does not utilize the \texttt{family-friendly} field. This will be corrected in section (b) of the graph.}
    \label{fig:ex_location_gen}
  \end{figure}
  
  \begin{figure}[ht]
    \centering
    \includegraphics[width=\linewidth]{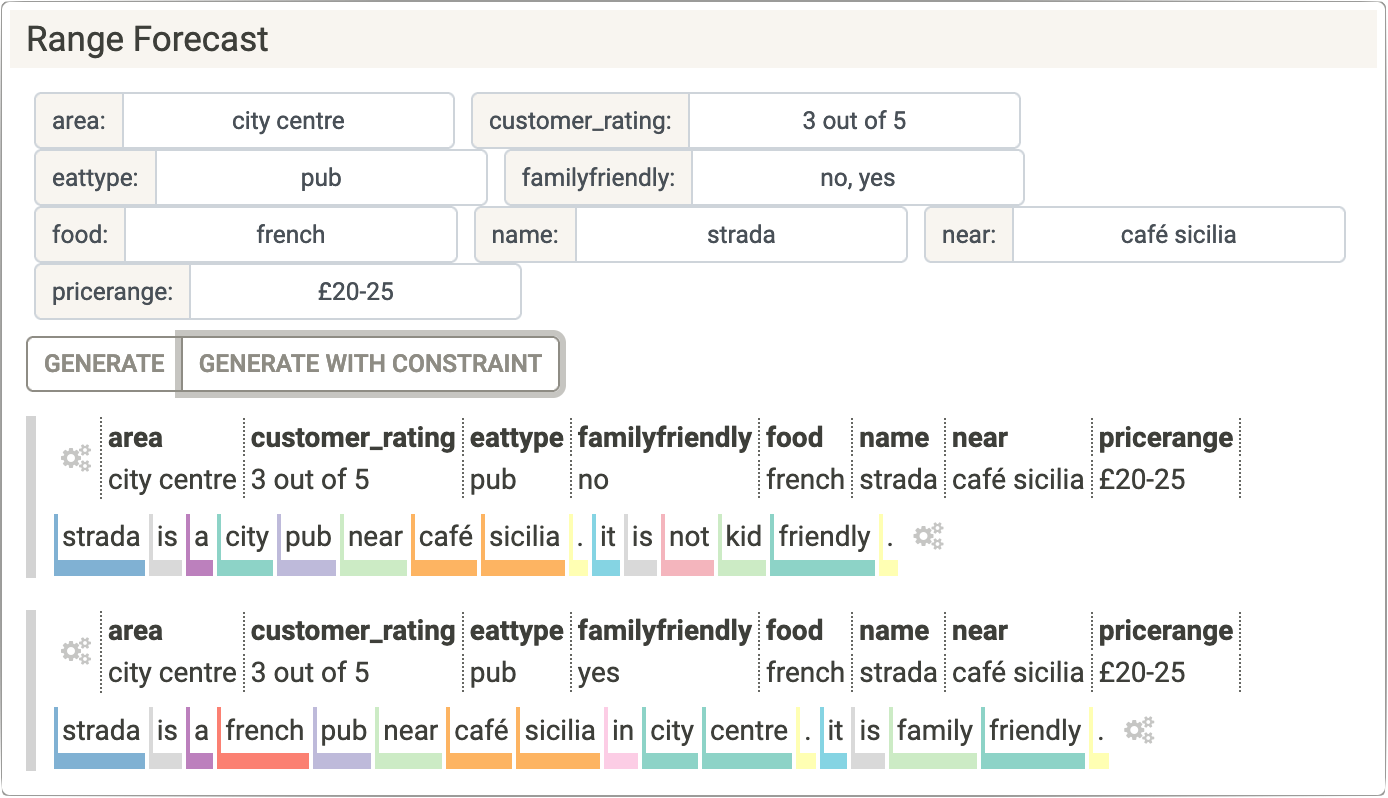}
    \caption{Outputs of Range Forecast for the constraint graph. By providing the range \texttt{no,yes} for \texttt{family-friendly}, the system generates multiple tables with different values and uses these to produce text under the constraint graph. The output shows reasonable results for both values. }
    \label{fig:ex_final}
  \end{figure}

Using the Forecast Generation component for free generation [T1], we can confirm from \autoref{fig:date_uc_all}a that the model has learned reasonable clusters for the control states.  We can see that the model uses red for year, purple for month, yellow for ordinal day, light blue for nominal day, and so on. This view confirms the model structure is correct and that the system will alternate between different styles of generation.

Furthermore, the Example Creation view in \autoref{fig:date_uc_all}b shows us that the control states inferred by the model [T2] also agree with the clustering observed. (Anecdotally, in early testing, the model architect  was able to find an error in the model implementation based on the interaction mode through this process.) 

Finally, the Constraint Graph view can be used to enforce that output text obeys a specified format. We consider the constraining to the format ``\texttt{the date is september the fourteenth, 2015 .}''. Note, that the yellow state has been edited to add a repeat loop allowing for variable length (in practice, one or two) day output text. See \autoref{fig:date_uc_all}c.

\noindent \textbf{Restaurant Recommendation}
For more complex use-cases, we turn to the problem of refining text for an assistive agent or bot. The E2E dataset \cite{novikova_e2e_2017} is a standard data set designed for benchmarking Table2Text generation systems to simulate conversational responses in a constrained environment. 
 Here the input $\vx$ is a table containing information about a restaurant (subset of eight different possible fields). The corresponding sentence $\vy$ is a description of the data table $\vx$. 
 
 For this use case, we assume the challenge is to constrain the output to highlight the \textit{location} of the restaurant and whether it is \textit{family-friendly}. There are several challenges that make this difficult.  In the beginning, we do not know the right alignments between the control states and the relevant table fields. We also need to allow for different possible orderings of these fields. Additionally, we do not know how fields like \textit{family-friendly} with \texttt{yes,no} values that cannot be copied directly are described.
 
 Let us start with the location of the restaurant. There are two table fields \texttt{area} and \texttt{near} related to location and we would like to include both of them. Using the \textit{Control States} (see \autoref{fig:ex_align}) section of the \textit{Generation Forecast} view we can determine how these fields are used. This section shows overall alignment between control states, table fields (inputs) and text (output). We can see that the model uses cyan (A) control states for \texttt{area}, outputting words such as \texttt{city} and \texttt{riverside} from this state, and orange (G) control states for \texttt{near}. 
  
We can now ensure that these fields are both used. We observe that location fields can appear in either order - (\texttt{area}, \texttt{near}) or (\texttt{near}, \texttt{area}). We encode this in a constraint graph shown in \autoref{fig:ex_location}a. The (a) section of this graph ensures that some descriptive text comes first (seed s), and then the model can fork to generate the location in either order. Upon constructing this section of the graph, we can check that this constrained generation is looking correct. The generation results in  \autoref{fig:ex_location_gen} show that the generation with constraints works as expected, producing initial text followed by full location descriptions such as \texttt{near raja indian cuisine in the city centre}.

However, this constraint graph does not yet ensure that the text mentions whether the restaurant is family-friendly. To first determine how the system might encode this property, we use the Manual Output tool in the Refine Constraints section. Ignoring the rest of the text, we manually type out the phrases \texttt{it is family friendly .} and \texttt{it is not family friendly .}. The system then finds the best control states for each of the input tokens. In this case, the system produces the same control states for \texttt{family friendly} but differs in the \texttt{not} state. We can then combine these two control paths using the Refine by Examples tool. This effectively creates an optional \textit{not} state in the graph, allowing both possibilities.

Finally, we combine this section (b) into the full constraint graph in ~\autoref{fig:ex_location}. Together this ensures we have both the location and the description of family friendliness. We can ensure this works using the \textit{Range Forecast} tool. This tool allows us to generate a range of possible input tables to test the output text. The outputs in \autoref{fig:ex_final} for \texttt{yes} and \texttt{no} values for the \texttt{family friendly} table field shows that our constraint works well in both cases and even generates text in a different manner (\texttt{kid friendly}) than our manual input.

\section{Implementation}
\label{sec:implementation}

GenNI requires interaction with a live model designed to facilitate interactive visualization and refinement. To support this, it uses tight integration of a model with the visual client. We based the interface between both parts on a REST API, and we used a custom generation system for the underlying model framework using Torch-Struct~\cite{rush_torch-struct_2020}.  We designed an API to allow easy access of free generations, controlled generation, inferred control, and interaction with the model's beam tree. Both the backend and frontend communicate constraints only through the $\vz$ control states that form the central white-box component of the model.

The model framework works within a FastAPI server to deliver content via a REST interface to the client. The client is written in Typescript. Most visualization components are using the d3js library. Source code, a demo instance, and a descriptive webpage are available at \url{https://genni.vizhub.ai}.

\section{Related Work}
\label{sec:related-work}

\subsection{Table2Text Generation Models}
\label{sec:relwork_tgm}

Methods for table2text generation are commonly divided into rule-based approaches, statistical methods, and neural models. Rule-based approaches merge domain knowledge into the text generation systems~\cite{moore_planning_1993, hovy_automated_1993, reiter_building_2000, belz_probabilistic_2007, bouayad-agha_content_2011}. The domain knowledge can be encoded using hand crafted templates that map the data directly to language~\cite{geldof_architecture_1997, theune_data_2001, mcroy_augmented_2003} or through rule-based transformations of semantic representations to produce the output text~\cite{reiter_building_1997, cahill_search_2000, reiter_nlg_1995, reiter_building_2000}. Some systems combine template-based methods with standard rule-based approaches \cite{busemann_flexible_1998}. 
Our system does not use manual rule-based approaches; however, the learned control states are reminiscent of templates since the codes learn to align with specific characteristics of the text output. 
In this way, our approach has some similarities to statistical approaches that learn rules from training data \cite{langkilde_generation_nodate, duboue_statistical_2003, howald_domain_2013}. 
Dou et al.~\cite{dou_data2text_2018} built a model called Data2Text Studio for automated text generation from structured data by extracting templates. Like our system, it provides the user tools to edit templates for models and APIs to generate text. However, our system uses a neural model and also constructs constraint graphs rather than hard-coded templates.

As with most tasks involving language modeling, neural network models have become popular in conditional text generation. These models have provided significant improvements in performance as compared to rule-based and statistical models. The most popular models are seq2seq models that use recurrent neural networks, especially LSTMs, \cite{sutskever_sequence_2014} and transformer-based models that replace recurrence with multi-headed attention in a feed-forward set-up~\cite{vaswani_attention_2017}. These seq2seq models have been used for conditional text generation by encoding the data as a source sequence and employing standard transduction methods \cite{kale_text--text_2020, mei_what_2016, dusek_sequence--sequence_2016, lebret_neural_2016}. Transformer-based models that have been trained on huge corpus of data \cite{radford_language_2019, devlin_bert_2019, kale_text--text_2020} such as GPT2 and BERT are commonly used to warm-start such models~\cite{raffel_exploring_2020, wolf_transfertransfo_2019}. Recently, transformer-based models similar to BERT~\cite{devlin_bert_2019} have been pre-trained on table dataset~ \cite{herzig_tapas_2020}. Our system uses recurrent network models for its different components. However, since the working of our system depends on the probabilistic generative model (similar to a statistical modeling approach) and not the underlying implementation, it should be able to leverage larger transformer-based models, pre-trained or not.

\subsection{Controllable Text Generation}

Standard neural network models trained end-to-end are black-box text generators, and it is difficult to control the generated text. To this extent, recently developed methods allow injecting control into these models. The controllable attributes can vary from topic, sentiment, politeness, tense, ordering of information, content, etc. These models learn control codes~\cite{keskar_ctrl_2019} that only moderate high-level attributes such as sentiment~\cite{hu_toward_2017, luo_learning_2019} and style~\cite{oraby_controlling_2018, shen_style_2017}, and thus can still generate text that differ at the word and phrase levels. Other models manipulate the syntactic structure of generated text \cite{chen_controllable_2019, iyyer_adversarial_2018, colin_generating_2018, deriu_syntactic_2018}.

For more fine-level properties, some models learn templates \cite{wiseman_learning_2018}, alignment between data and text \cite{chan_cocon_2020, shen_neural_2020}. Our system, which is built upon the model proposed by Li et al. \cite{li_posterior_2020} with a linear chain conditional random field in the inference network and trained using Gumbel approximation following Fu et al. \cite{fu_latent_2020} is most similar to these approaches.  The control states learned to control some high-level semantics of the words generated and can be used to extract templates (in the form of constraint graphs). The semi-supervised training done for posterior regularization performs soft alignment between the text and the data.

\subsection{Interactive Interfaces for Text Generation}
\label{sec:relwork_textgen}

Interactive interfaces for free text generation are increasingly popular. ``Write with transformer''~\cite{huggingface_write_2021} completes paragraphs that have been started by user input using transformer models like GPT-2. Some commercial applications like GMail use language models to improve their sentence completion. The Google Translate UI uses text generation for translation. TabNine~\cite{codota_dot_com_ltd_tabnine_nodate} offers language generation for programming languages integrated into multiple IDEs. MixingBoard~\cite{gao_mixingboard_2020} demonstrates interfaces for knowledge grounded stylized text generation. Text generation models can also be used to detect if the models have created an input text themselves~\cite{gehrmann_gltr_2019, zellers_defending_2020}. Note, though, that these differ from systems that focus on conditional generation. 

CSI:Summary~\cite{gehrmann_visual_2020} describes a system for text summarization that uses a controlled generation model. Outputs can be constrained as a response to user interactions. Data2Text Studio~\cite{dou_data2text_2018} allows formulation of constraints as set of Boolean rules.  GenNI builds on the work of CSI:Summary.

\subsection{Explainable AI for Sequence Models}
\label{sec:relwork_xai}
Visualization for explainable AI is a very active research topic resulting in high-frequent publications. Hohman et al.~\cite{hohman_visual_2019} provide a comprehensive start into this topic. Here, we exclusively focus on approaches for sequence models.

As the fundamental and earliest building block, RNNs have been the subject of study. The ``unreasonable'' effectiveness of RNNs for encoding sequential information \cite{karpathy_visualizing_2015} can be interactively explored by approaches like LSTMVis~\cite{strobelt_lstmvis_2018}, RNNVis~\cite{ming_understanding_2017} or ProtoSteer~\cite{ming_protosteer_2020}. Several methods utilize the model's gradient and map them to model input for analysis~\cite{cashman_rnnbow_2018, madsen_visualizing_2019}.

Current state-of-the-art deep learning NLP methods, like seq2seq models or transformers, are more complex and require interactive methods to investigate this complexity. Seq2SeqVis~\cite{strobelt_seq2seq-vis_2019} enables interactive what-if analysis of the five parts of a seq2seq model. BertViz\cite{vig_multiscale_2019} and exBert\cite{hoover_exbert_2020} allow a deep look into the attention mechanisms of transformer models. These tools provide interactivity to analyze single examples. GenNI extends from this idea and aims to generalize from concrete examples to a set of applicable rules for the whole dataset.

\section{Conclusions and Future Work}
\label{sec:conclusion}
We present GenNI, a system for collaborative development of data-backed text generation systems.
Unlike many systems developed for understanding deep learning models for NLP, GenNI is designed to
help users produce actionable constraints that can be used with systems designed for user control.
The system facilitates a collaborative interaction with users refining explicit constraints on the model
and the AI system forecasting generations on new data.

This style of controllable model can be designed for many different tasks in NLP and related domains.
Building models with user understandable controls opens up the ability for explicitly collaborative systems
as opposed to the trade-off of rule-based systems and full AI-driven outputs. Visual interaction plays a key
role for making it possible for a user to intuit, develop, and apply these constraints in a test environment
as well as deploy them in real systems. In GenNI, the encodings and structure used were targeted specifically to a
class of controllable generation models, but the approach of a single control state shared in a refine/forecast setting
can be applied much more broadly. Future work will look to develop shared encodings that can be applied to a wide class
of controllable NLP models.

\section{Acknowledgements}
\label{sec:acknowledgement}

This work was partially supported by NSF grant III-1901030 and a Google Faculty Research Award.

\bibliographystyle{abbrv-doi}

\bibliography{template}
\end{document}